\documentclass[11pt,a4paper]{article}
\usepackage[hyperref]{emnlp2018}
\usepackage{times}
\usepackage{latexsym}
\usepackage{graphicx}
\usepackage{url}
\usepackage{multirow}
\usepackage{caption}
\usepackage{dblfloatfix}
\usepackage{booktabs}
\usepackage{amsmath}
\usepackage{comment}
\usepackage{todonotes}
\usepackage{footnote}
\usepackage{footmisc}

\aclfinalcopy % Uncomment this line for the final submission
\def\aclpaperid{71} %  Enter the acl Paper ID here

\newcommand{\astfootnote}[1]{
\let\oldthefootnote=\thefootnote
\setcounter{footnote}{0}
\renewcommand{\thefootnote}{\fnsymbol{footnote}}
\footnote{#1}
\let\thefootnote=\oldthefootnote
}

\usepackage{lipsum}

\newcommand\blfootnote[1]{%
  \begingroup
  \renewcommand\thefootnote{}\footnote{#1}%
  \addtocounter{footnote}{-1}%
  \endgroup
}

\title{MultiWOZ - A Large-Scale Multi-Domain Wizard-of-Oz Dataset for Task-Oriented Dialogue Modelling} 
\author{Pawe\l~Budzianowski{$^1$}, Tsung-Hsien~Wen{$^{2*}$}, Bo-Hsiang Tseng{$^1$}, \\   \textbf{I\~nigo~Casanueva{$^{2*}$}, Stefan~Ultes{$^1$}, Osman Ramadan{$^1$} and Milica~Ga{\v s}i{\' c}}{$^1$}\\ 
{$^1$}Department of Engineering, University of Cambridge, UK, \\
{$^2$}PolyAI, London, UK\\
 {\tt \{pfb30,mg436\}@cam.ac.uk}\\
}
\date{}
\begin{document}
\maketitle

\begin{abstract}
\renewcommand{\thefootnote}{\fnsymbol{footnote}} 
\blfootnote{$^{*}$The work was done while at the University of Cambridge.}

Even though machine learning has become the major scene in dialogue research community, the real breakthrough has been blocked by the scale of data available.
To address this fundamental obstacle, we introduce the Multi-Domain Wizard-of-Oz dataset (MultiWOZ), a fully-labeled collection of human-human written conversations spanning over multiple domains and topics.
At a size of $10$k dialogues, it is at least one order of magnitude larger than all previous annotated task-oriented corpora.
The contribution of this work apart from the open-sourced dataset labelled with dialogue belief states and dialogue actions is two-fold:
firstly, a detailed description of the data collection procedure along with a summary of data structure and analysis is provided. The proposed data-collection pipeline is entirely based on crowd-sourcing without the need of hiring professional annotators;
secondly, a set of benchmark results of belief tracking, dialogue act and response generation is reported, which shows the usability of the data and sets a baseline for future studies.

\end{abstract}

\section{Introduction}
Conversational Artificial Intelligence (Conversational AI) is one of the long-standing challenges in computer science and artificial intelligence since the Dartmouth Proposal~\cite{dartmouth1955}.
As human conversation is inherently complex and ambiguous, learning an open-domain conversational AI that can carry on arbitrary tasks is still very far-off~\citep{vinyals2015neural}.
As a consequence, instead of focusing on creating ambitious conversational agents that can reach human-level intelligence, industrial practice has focused on building task-oriented dialogue systems~\cite{POMDP-review} that can help with specific tasks such as flight reservation~\cite{Seneff2000} or bus information~\cite{raux2005let}.
As the need of hands-free use cases continues to grow, building a conversational agent that can handle tasks across different application domains has become more and more prominent~\cite{ram2018conversational}.

Dialogues systems are inherently hard to build because there are several layers of complexity: the noise and uncertainty in speech recognition~\cite{black2011spoken}; the ambiguity when understanding human language~\cite{williams2013dialog}; the need to integrate third-party services and dialogue context in the decision-making~\cite{traum2003information,paek2008automating}; and finally, the ability to generate natural and engaging responses~\cite{stent2005evaluating}.
These difficulties have led to the same solution of using statistical framework and machine learning for various system components, such as natural language understanding~\cite{henderson2013deep,mesnil2015using,mrkvsic2017neural}, dialogue management~\cite{GPRL,tegho2018benchmarking}, language generation~\cite{wensclstm15,kiddon2016globally}, and even end-to-end dialogue modelling~\cite{zhao2016towards,wen2016network,eric2017key}.

\begin{table*}[t]
\begin{center}
\resizebox{\textwidth}{!}{%
    \begin{tabular}{lccccccc}
    \toprule
    \textbf{Metric} & \textbf{DSTC2} & \textbf{SFX} &  \textbf{WOZ2.0} &\textbf{FRAMES}  & \textbf{KVRET}  & \textbf{M2M} & \textbf{MultiWOZ} \\
    \midrule
   \# Dialogues & 1,612 & 1,006 & 600 & 1,369 & 2,425 & 1,500 & \textbf{8,438} \\
    Total \# turns & 23,354 & 12,396 &  4,472 & 19,986 & 12,732 & 14,796 & \textbf{113, 556} \\
    Total \# tokens & 199,431 & 108,975 &50,264
 & 251,867 & 102,077 & 121,977 & \textbf{1,490,615} \\
    Avg. turns per dialogue & 14.49 &12.32&  7.45 & \textbf{14.60} & 5.25  & 9.86  & 13.46\\
    Avg. tokens per turn & 8.54  & 8.79& 11.24 & 12.60 & 8.02  & 8.24  & \textbf{13.13} \\
  
   Total unique tokens & 986  & 1,473 & 2,142 & 12,043 & 2,842 & 1,008 & \textbf{23689} \\
  \# Slots & 8 & 14 & 4 & \textbf{61} & 13 & 14 & 24 \\
    \# Values & 212 & 1847& 99 &3871 &1363& 138 & \textbf{4510} \\
    \bottomrule
    \end{tabular}}
\end{center}
\caption{\label{fig:corpus_comparison} Comparison of %the WOZ3
our corpus to similar data sets. Numbers in bold indicate best value for the respective metric. The numbers are provided for the training part of data except for FRAMES data-set were such division was not defined.}
%\vspace*{-1em}
\end{table*}

To drive the progress of building dialogue systems using data-driven approaches, a number of conversational corpora have been released in the past.
Based on whether a structured annotation scheme is used to label the semantics, these corpora can be roughly divided into two categories: corpora with structured semantic labels~\cite{hemphill1990atis,williams2013dialog,asri2017frames,wen2016network,eric2017key,shah2018building}; and corpora without semantic labels but with an implicit user goal in mind~\cite{ritter2010unsupervised,lowe2015ubuntu}.
Despite these efforts, aforementioned datasets are usually constrained in one or more dimensions such as missing proper annotations, only available in a limited capacity, lacking multi-domain use cases, or having a negligible linguistic variability.

This paper introduces the Multi-Domain Wizard-of-Oz (MultiWOZ) dataset, a large-scale multi-turn conversational corpus with dialogues spanning across several domains and topics.
Each dialogue is annotated with a sequence of dialogue states and corresponding system dialogue acts~\cite{Traum1999}.
Hence, MultiWOZ can be used to develop individual system modules as separate classification tasks and serve as a benchmark for existing modular-based approaches.
On the other hand, MultiWOZ has around $10$k dialogues, which is at least one order of magnitude larger than any structured corpus currently available.
This significant size of the corpus allows researchers to carry on end-to-end based dialogue modelling experiments, which may facilitate a lot of exciting ongoing research in the area.

This work presents the data collection approach, a summary of the data structure, as well as a series of analyses of the data statistics. To show the potential and usefulness of the proposed MultiWOZ corpus, benchmarking baselines of belief tracking, natural language generation and end-to-end response generation have been conducted and reported. The dataset and baseline models will be freely available online.\footnote{\url{https://github.com/budzianowski/multiwoz}}

\section{Related Work}
Existing datasets can be roughly grouped into three categories: machine-to-machine, human-to-machine, and human-to-human conversations. A detailed review of these categories is presented below.

{\noindent \bf Machine-to-Machine} \hspace{1.5mm}
Creating an environment with a simulated user enables to exhaustively generate dialogue templates. These templates can be mapped to a natural language by either pre-defined rules \cite{bordes2016learning} or crowd workers \cite{shah2018building}. Such approach ensures a diversity and full coverage of all possible dialogue outcomes within a certain domain. However, the naturalness of the dialogue flows relies entirely on the engineered set-up of the user and system bots. This poses a risk of a mismatch between training data and real interactions harming the interaction quality. 
Moreover, these datasets do not take into account noisy conditions often experienced in real interactions \cite{black2011spoken}.\\

{\noindent \bf Human-to-Machine} \hspace{1.5mm}
Since collecting dialogue corpus for a task-specific application from scratch is difficult, most of the task-oriented dialogue corpora are fostered based on an existing dialogue system. One famous example of this kind is the Let's Go Bus Information System which offers live bus schedule information over the phone \cite{raux2005let} leading to the
first Dialogue State Tracking Challenge \cite{williams2013dialog}.
Taking the idea of the Let's Go system forward, the second and third DSTCs~\cite{Henderson2014b,henderson2014third} have produced bootstrapped human-machine datasets for a restaurant search domain in the Cambridge area, UK.
Since then, DSTCs have become one of the central research topics in the dialogue community~\cite{kim2016fifth,kim2017fourth}.

While human-to-machine data collection is an obvious solution for dialogue system development, it is only possible with a provision of an existing working system.
Therefore, this chicken (system)-and-egg (data) problem limits the use of this type of data collection to existing  system improvement instead of developing systems in a completely new domain.
What is even worse is that the capability of the initial system introduces additional biases to the collected data, which may result in a mismatch between the training and testing sets~\cite{wen2016multi}. The limited understanding capability of the initial system may prompt the users to adapt to simpler input examples that the system can understand but are not necessarily natural in conversations.\\

{\noindent \bf Human-to-Human} \hspace{1.5mm}
Arguably, the best strategy to build a natural conversational system may be to have a system that can directly mimic human behaviors through learning from a large amount of real human-human conversations.
With this idea in mind, several large-scale dialogue corpora have been released in the past, such as the Twitter~\cite{ritter2010unsupervised} dataset, the Reddit conversations~\cite{schrading2015analysis}, and the Ubuntu technical support corpus~\cite{lowe2015ubuntu}.
Although previous work~\cite{vinyals2015neural} has shown that a large learning system can learn to generate interesting responses from these corpora, the lack of grounding conversations onto an existing knowledge base or APIs limits the usability of developed systems.
Due to the lack of an explicit goal in the conversation, recent studies have shown that systems trained with this type of corpus not only struggle in generating consistent and diverse responses~\cite{LiGBGD15} but are also extremely hard to evaluate~\cite{liu2016not}.\\

%{\noindent \bf Wizard-of-OZ} \hspace{1.5mm}
In this paper, we focus on a particular type of human-to-human data collection. The Wizard-of-Oz framework (WOZ)~\cite{kelley1984iterative} was first proposed as an iterative approach to improve user experiences when designing a conversational system.
The goal of WOZ data collection is to log down the conversation for future system development.
One of the earliest dataset collected in this fashion is the ATIS corpus~\cite{hemphill1990atis}, where conversations between a client and an airline help-desk operator were recorded.

More recently, \citet{wen2016network} have shown that the WOZ approach can be applied to collect high-quality typed conversations where a machine learning-based system can learn from.
By modifying the original WOZ framework to make it suitable for crowd-sourcing, a total of $676$ dialogues was collected via Amazon Mechanical Turk. The corpus was later extended to additional two languages for cross-lingual research~\cite{mrkvsic2017semantic}.
Subsequently, this approach is followed by~\citet{asri2017frames} to collect the Frame corpus in a more complex travel booking domain, and \citet{eric2017key} to collect a corpus of conversations for in-car navigation.
Despite the fact that all these datasets contain highly natural conversations comparing to other human-machine collected datasets, they are usually small in size with only a limited domain coverage.

 \begin{figure}[t!]
\centering
 \includegraphics[width=.85\linewidth]{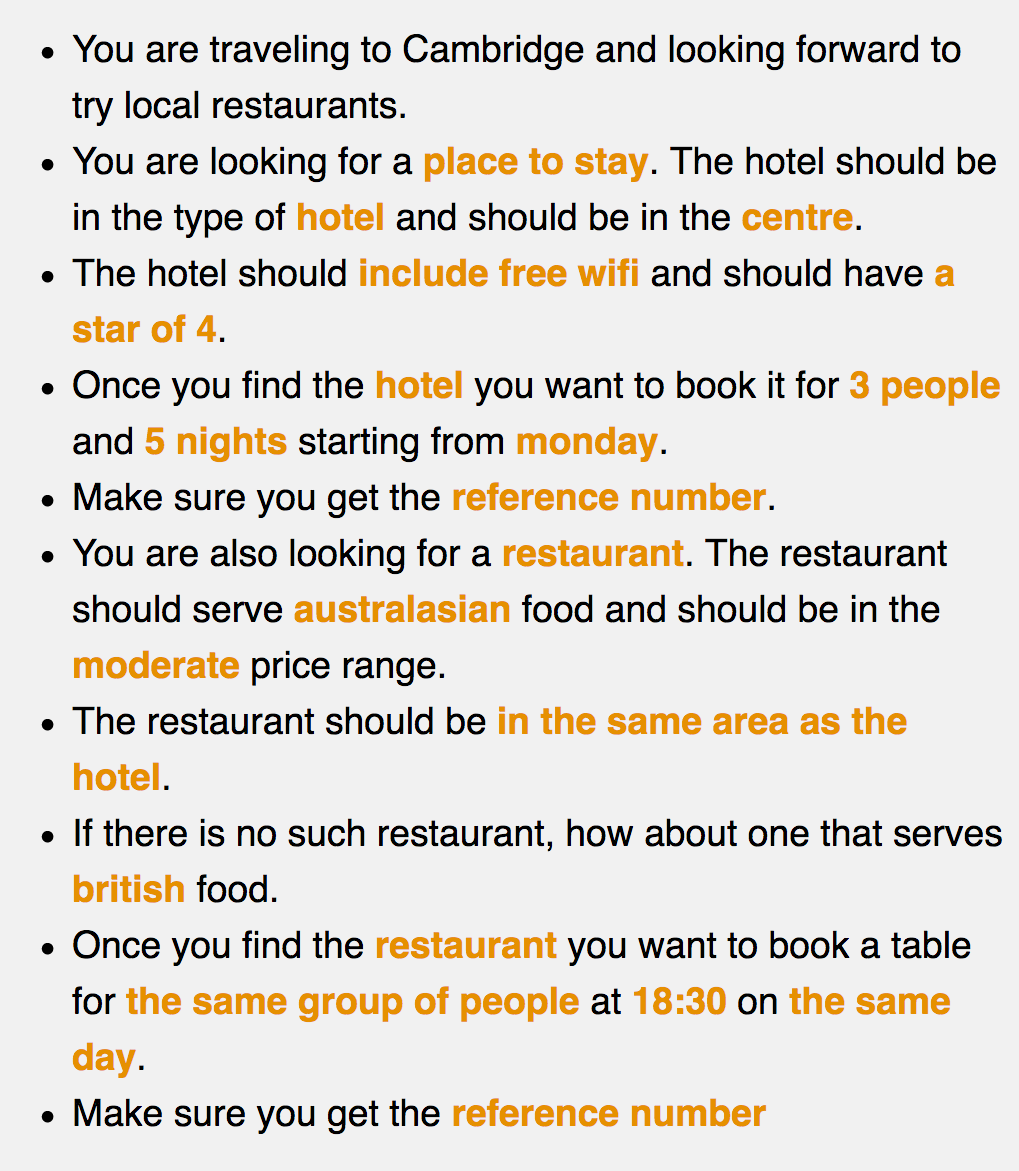}
  \caption{A sample task template spanning over three domains - hotels, restaurants and booking.}
  \label{fig:goal}
 % \vspace*{-1em}
\end{figure}

\begin{table*}[t!]
\centering
\caption{Full ontology for all domains in our data-set.
The upper script indicates which domains it belongs to. *: universal, 1: restaurant, 2: hotel, 3: attraction, 4: taxi, 5: train, 6: hospital, 7: police. \label{tab:ontology}}
\begin{tabular}{|l|l|}
\hline
act type & \begin{tabular}[c]{@{}l@{}}$\text{inform}^{*}$ / $\text{request}^{*}$ / $\text{select}^{123}$ / $\text{recommend/}^{123}$ / $\text{not\ found}^{123}$ \\ $\text{request\ booking\ info}^{123}$ / $\text{offer\ booking}^{1235}$ / $\text{inform\ booked}^{1235}$ / $\text{decline\ booking}^{1235}$ \\
$\text{welcome}^{*}$ /$\text{greet}^{*}$ / $\text{bye}^{*}$  / $\text{reqmore}^{*}$  \end{tabular}                                                                                                                                      \\ \hline
slots    & \begin{tabular}[c]{@{}l@{}}$\text{address}^{*}$ / $\text{postcode}^{*}$ / $\text{phone}^{*}$ / $\text{name}^{1234}$ / $\text{no\ of\ choices}^{1235}$ / $\text{area}^{123}$ /\\ $\text{pricerange}^{123}$ / $\text{type}^{123}$ / $\text{internet}^{2}$ / $\text{parking}^{2}$ / $\text{stars}^{2}$ / $\text{open\ hours}^{3}$ / $\text{departure}^{45}$\\ $\text{destination}^{45}$ / $\text{leave\ after}^{45}$ / $\text{arrive\ by}^{45}$ / $\text{no\ of\ people}^{1235}$ / $\text{reference no.}^{1235}$ /\\ $\text{trainID}^{5}$ / $\text{ticket\ price}^{5}$ / $\text{travel\ time}^{5}$ / $\text{department}^{7}$ / $\text{day}^{1235}$ / $\text{no\ of\ days}^{123}$
\end{tabular} \\ \hline
\end{tabular}
%\vspace*{-1em}
  \label{tab:ontology}
\end{table*}

\section{Data Collection Set-up}
Following the Wizard-of-Oz set-up \cite{kelley1984iterative}, corpora of annotated dialogues can be gathered at relatively low costs and with a small time effort. This is in contrast to previous approaches~\cite{Henderson2014a} and such WOZ set-up has been successfully validated by \citet{wen2016network} and \citet{asri2017frames}.

Therefore, we follow the same process to create a large-scale corpus of natural human-human conversations. Our goal was to collect multi-domain dialogues. To overcome the need of relying the data collection to a small set of trusted workers\footnote{Excluding annotation phase.}, the collection set-up was designed to provide an easy-to-operate system interface for the Wizards and easy-to-follow goals for the users. This resulted in a bigger diversity and semantical richness of the collected data (see Section \ref{sec:analysis}). Moreover, having a large set of workers mitigates the problem of artificial encouragement of a variety of behavior from users. A detailed explanation of the data-gathering process from both sides is provided below. Subsequently, we show how the crowd-sourcing scheme can also be employed to annotate the collected dialogues with dialogue acts.

\subsection{Dialogue Task}
\label{sec:task}
The domain of a task-oriented dialogue system is often defined by an ontology, a structured representation of the back-end database. The ontology defines all entity attributes called slots and all possible values for each slot.
In general, the slots may be divided into \emph{informable} slots and \emph{requestable} slots. \emph{Informable} slots are attributes that allow the user to constrain the search (e.g., area or price range). \emph{Requestable} slots represent additional information the users can request about a given entity (e.g., phone number).
Based on a given ontology spanning several domains, a task template was created for each task through random sampling. This results in single and multi-domain dialogue scenarios and domain specific constraints were generated. In domains that allowed for that, an additional booking requirement was sampled with some probability.

To model more realistic conversations, goal changes are encouraged. With a certain probability, the initial constraints of a task may be set to values so that no matching database entry exists. Once informed about that situation by the system, the users only needed to follow the goal which provided alternative values.

\subsection{User Side}
To provide information to the users, each task template is mapped to natural language. Using heuristic rules, the task is then gradually introduced to the user to prevent an overflow of information. The goal description presented to the user is dependent on the number of turns already performed. Moreover, if the user is required to perform a sub-task (for example - booking a venue), these sub-goals are shown straight-away along with the main goal in the given domain.
This makes the dialogues more similar to spoken conversations.\footnote{However, the length of turns are significantly longer than with spoken interaction (Section \ref{sec:analysis}).}
Figure \ref{fig:goal} shows a sampled task description spanning over two domains with booking requirement. Natural incorporation of co-referencing and lexical entailment into the dialogue was achieved through implicit mentioning of some slots in the goal. 

\begin{figure*}[t!]
\centering
\begin{minipage}{.48\textwidth}
  \centering
  \includegraphics[width=1.\linewidth]{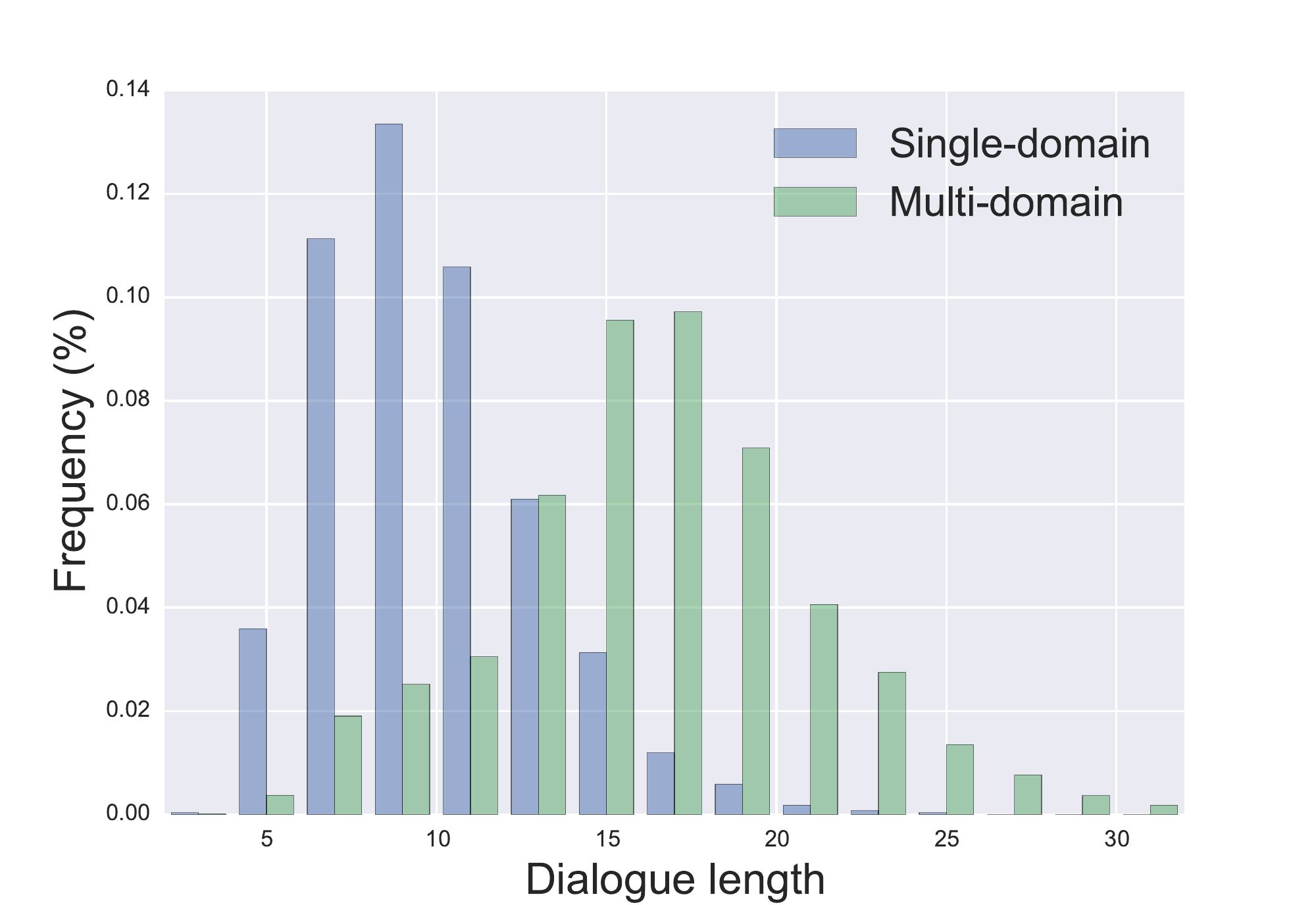}
  %\captionof{figure}{}
  \label{fig:test1}
  % \vspace*{-1em}
\end{minipage}\hfill
\begin{minipage}{.48\textwidth}
  \centering
  \includegraphics[width=1.\linewidth]{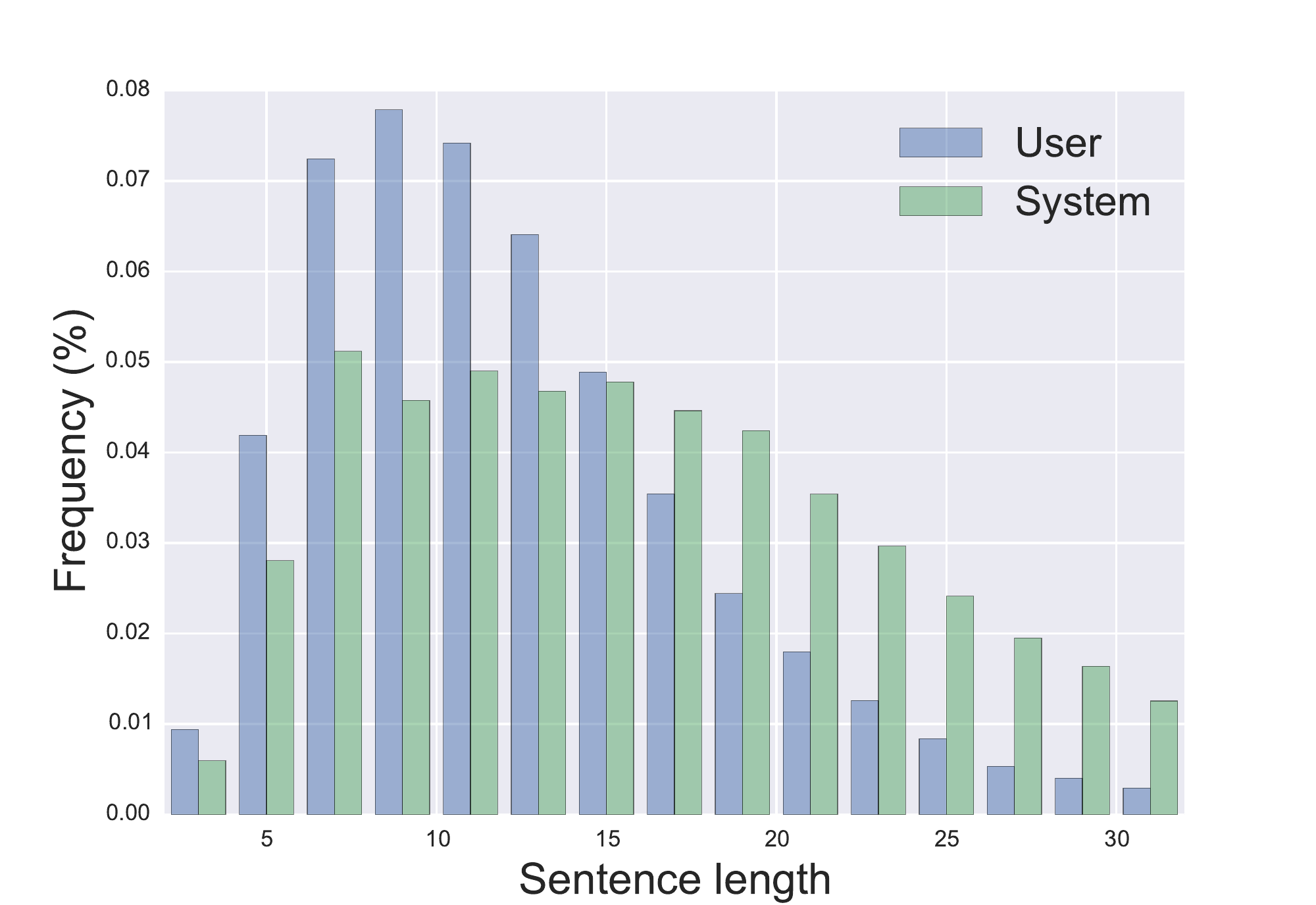}
  %\captionof{figure}{Number of dialogue acts per turn.}
  \label{fig:test2}
  %\vspace*{-1em}
\end{minipage}
%\vspace*{-1em}
\caption{Dialogue length distribution (left) and distribution of number of tokens per turn (right).} \label{fig:length_stats}
%\vspace*{-1em}
\end{figure*}

\subsection{System Side}\label{sec:sysside}
The wizard is asked to perform a role of a clerk by providing information required by the user. He is given an easy-to-operate graphical user interface to the back-end database. The wizard conveys the information provided by the current user input through a web form. This information is persistent across turns and is used to query the database. Thus, the annotation of a belief state is performed implicitly while the wizard is allowed to fully focus on providing the required information. Given the result of the query (a list of entities satisfying current constraints), the wizard either requests more details or provides the user with the adequate information. At each system turn, the wizard starts with the results of the query from the previous turn.

To ensure coherence and consistency, the wizard and the user alike first need to go through the dialogue history to establish the respective context. We found that even though multiple workers contributed to one dialogue, only a small margin of dialogues were incoherent.

\subsection{Annotation of Dialogue Acts}
Arguably, the most challenging and time-consuming part of any dialogue data collection is  the process of annotating dialogue acts. One of the major challenges of this task is the definition of a set and structure of dialogue acts \cite{traum1992conversation, bunt2006dimensions}. In general, a dialogue act consists of the intent (such as request or inform) and slot-value pairs. For example, the act {\verb|inform(domain=hotel,price=expensive)|}
has the intent \textit{inform}, where the user is informing the system to constrain the search to expensive hotels.

Expecting a big discrepancy in annotations between annotators, we initially ran three trial tests over a subset of dialogues using Amazon Mechanical Turk. Three annotations per dialogue were gathered resulting in around $750$ turns. As this requires a multi-annotator metric over a multi-label task, we used Fleiss' kappa metric \cite{fleiss1971measuring} per single dialogue act. Although the weighted kappa value averaged over dialogue acts was at a high level of $0.704$, we have observed many cases of very poor annotations and an unsatisfactory coverage of dialogue acts.
Initial errors in annotations and suggestions from crowd workers gradually helped us to expand and improve the final set of dialogue acts from $8$ to $13$ - see Table \ref{tab:ontology}. 

The variation in annotations made us change the initial approach. We ran a two-phase trial to first identify set of workers that perform well. Turkers were asked to annotate an illustrative, long dialogue which covered many problematic examples that we have observed in the initial run described above. All submissions that were of high quality were inspected and corrections were reported to annotators. Workers were asked to re-run a new trial dialogue. Having passed the second test, they were allowed to start annotating real dialogues. This procedure resulted in a restricted set of annotators performing high quality annotations. Appendix \ref{sec:website} contains a demonstration of a created system. 

\subsection{Data Quality}
Data collection was performed in a two-step process. First, all dialogues were collected and then the annotation process was launched. This setup allowed the dialogue act annotators to also report errors (e.g., not following the task or confusing utterances) found in the collected dialogues. As a result, many errors could be corrected. Finally, additional tests were performed to ensure that the provided information in the dialogues match the pre-defined goals.

To estimate the inter-annotator agreement, the averaged weighted kappa value for all dialogue acts was computed over $291$ turns. With $\kappa = 0.884$, an improvement in agreement between annotators was achieved although the size of action set was significantly larger.

\begin{figure*}[t!]
\centering
\begin{minipage}{.5\textwidth}
  \centering
  \includegraphics[width=1.\linewidth]{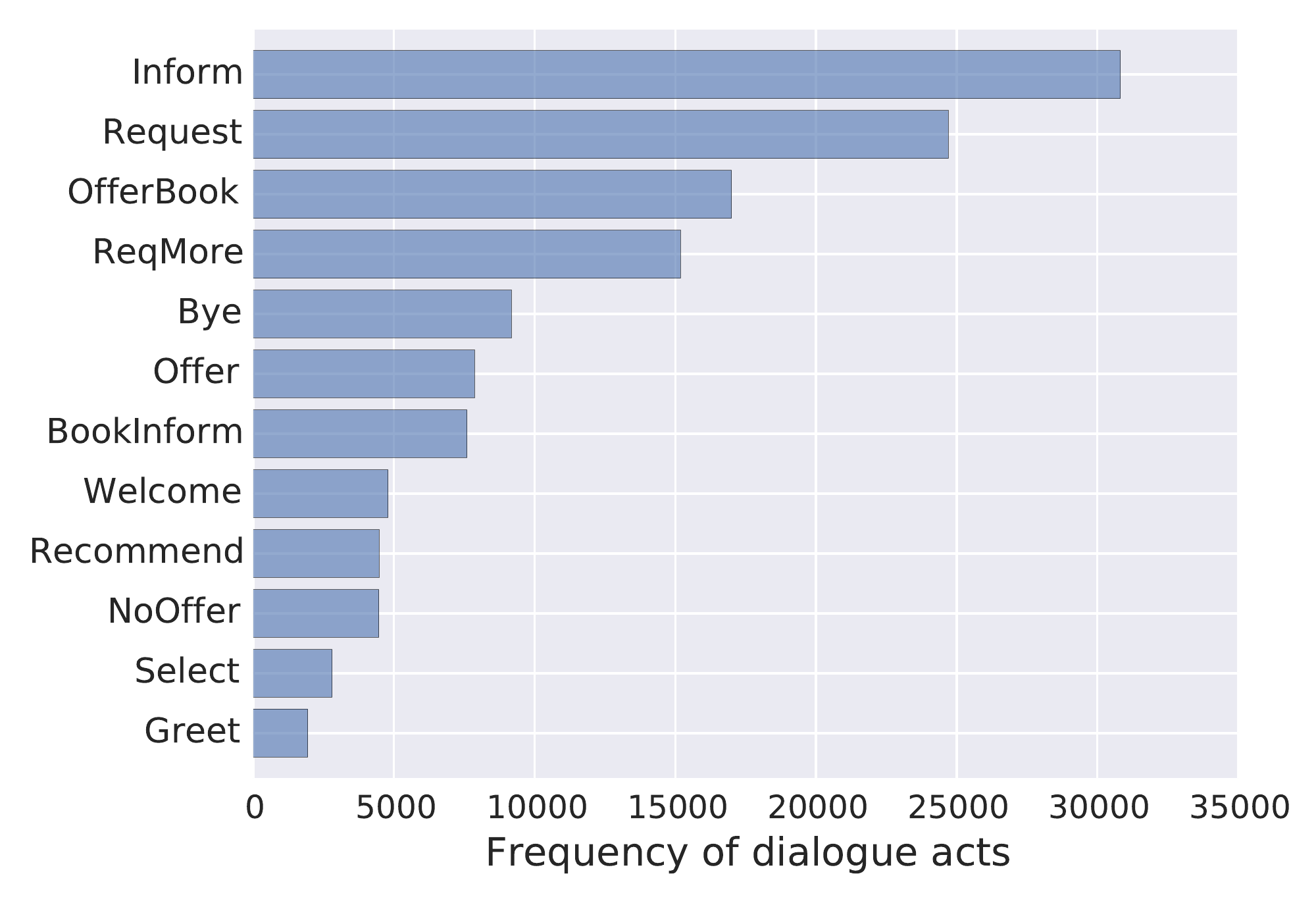}
  %\captionof{figure}{}
  \label{fig:test1}
\end{minipage}%
\begin{minipage}{.5\textwidth}
  \centering
  \includegraphics[width=.95\linewidth]{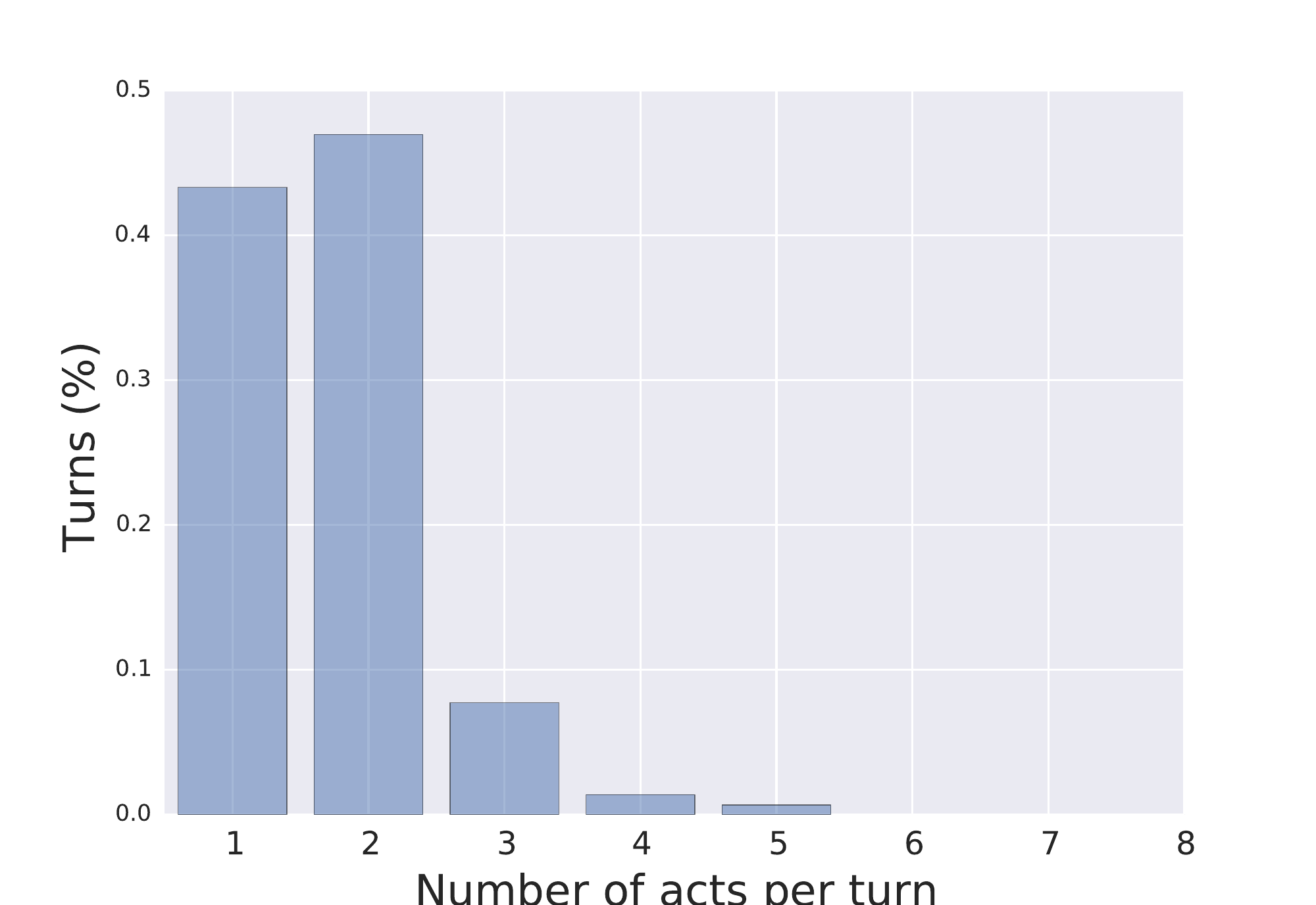}
  %\captionof{figure}{Number of dialogue acts per turn.}
  \label{fig:test2}
  \vspace*{2.5em}
\end{minipage}
%\vspace*{-1em}
\caption{Dialogue acts frequency (left) and number of dialogue acts per turn (right) in the collected corpus.} \label{fig:acts_fig}
%\vspace*{-1em}
\end{figure*}

\section{MultiWOZ Dialogue Corpus}
The main goal of the data collection was to acquire highly natural conversations between a tourist and a clerk from an information center in a touristic city. We considered various possible dialogue scenarios ranging from requesting basic information about attractions through booking a hotel room or travelling between cities. In total, the presented corpus consists of $7$ domains - \emph{Attraction, Hospital, Police, Hotel, Restaurant, Taxi, Train}. The latter four are extended domains which include the sub-task \emph{Booking}. Through a task sampling procedure (Section \ref{sec:task}), the dialogues cover between $1$ and $5$ domains per dialogue thus greatly varying in length and complexity. This broad range of domains allows to create scenarios where domains are naturally connected. For example, a tourist needs to find a hotel, to get the list of attractions and to book a taxi to travel between both places.
Table \ref{tab:ontology} presents the global ontology with the list of considered dialogue acts.   

\subsection{Data Statistics}
Following data collection process from the previous section, a total of $10,438$ dialogues were collected. Figure \ref{fig:length_stats} (left) shows the dialogue length distribution grouped by single and multi domain dialogues. Around $70\%$ of dialogues have more than $10$ turns which shows the complexity of the corpus. The average number of turns are $8.93$ and $15.39$ for single and multi-domain dialogues respectively with $115,434$ turns in total.
Figure \ref{fig:length_stats} (right) presents a distribution over the turn lengths. As expected, the wizard replies are much longer - the average sentence lengths are $11.75$ and $15.12$ for users and wizards respectively. The responses are also more diverse thus enabling the training of more complex generation models.

Figure \ref{fig:acts_fig} (left) shows the distribution of dialogue acts annotated in the corpus. We present here a summarized list where different types of actions like \emph{inform} are grouped together. The right graph in the Figure \ref{fig:acts_fig} presents the distribution of number of acts per turn. Almost $60\%$ of dialogues turns have more than one dialogue act showing again the richness of system utterances. These create a new challenge for reinforcement learning-based models requiring them to operate on concurrent actions.
 
In total, $1,249$ workers contributed to the corpus creation with only few instances of intentional wrongdoing. Additional restrictions were added to automatically discover instances of very short utterances, short dialogues or missing single turns during annotations. All such cases were corrected or deleted from the corpus.

\subsection{Data Structure}
There are $3,406$ single-domain dialogues that include booking if the domain allows for that and $7,032$ multi-domain dialogues consisting of at least $2$ up to $5$ domains. To enforce reproducibility of results, the corpus was randomly split into a train, test and development set. The test and development sets contain $1$k examples each. Even though all dialogues are coherent, some of them were not finished in terms of task description. Therefore, the validation and test sets only contain fully successful dialogues thus enabling a fair comparison of models. 

Each dialogue consists of a goal, multiple user and system utterances as well as a belief state  and set of dialogue acts with slots per turn. Additionally, the task description in natural language presented to turkers working from the visitor's side is added.

\subsection{Comparison to Other Structured Corpora}\label{sec:analysis}
To illustrate the contribution of the new corpus, we compare it on several important statistics with the DSTC2 corpus \cite{Henderson2014a}, the SFX corpus \cite{gavsic2014incremental}, the WOZ$2.0$ corpus \cite{wen2016network}, the FRAMES corpus \cite{asri2017frames}, the KVRET corpus \cite{eric2017key}, and the M2M corpus \cite{shah2018building}. Figure~\ref{fig:corpus_comparison} clearly shows that our corpus compares favorably to all other data sets on most of the metrics with the number of total dialogues, the average number of tokens per turn and the total number of unique tokens as the most prominent ones. Especially the latter is important as it is directly linked to linguistic richness.

\section{MultiWOZ as a New Benchmark}
The complexity and the rich linguistic variation in the collected MultiWOZ dataset makes it a great benchmark for a range of dialogue tasks. To show the potential usefulness of the MultiWOZ corpus, we break down the dialogue modelling task into three sub-tasks and report a benchmark result for each of them: dialogue state tracking, dialogue-act-to-text generation, and dialogue-context-to-text generation. These results illustrate new challenges introduced by the MultiWOZ dataset for different dialogue modelling problems.
%\cite multi domain and  lidm
\subsection{Dialogue State Tracking}
\label{sec:nlu}
A robust natural language understanding and dialogue state tracking is the first step towards building a good conversational system.
Since multi-domain dialogue state tracking is still in its infancy and there are not many comparable approaches available \cite{rast:17}, we instead report our state-of-the-art result on the restaurant subset of the MultiWOZ corpus as the reference baseline. The proposed method~\cite{ramadan2018large} exploits the semantic similarity between dialogue utterances and the ontology terms which allows the information to be shared across domains. Furthermore, the model parameters are independent of the ontology and belief states, therefore the number of the parameters does not increase with the size of the domain itself.\footnote{The model is publicly available at \url{https://github.com/osmanio2/multi-domain-belief-tracking}} 

\begin{table}[h!]
\centering
\begin{tabular}{ |c|c|c|}
\hline
\multirow{ 2}{*}{\textbf{Slot}} & \multirow{ 2}{*}{\textbf{WOZ 2.0}} & \textbf{MultiWOZ}  \\
& & \textbf{(restaurant)}\\
\hline
Overall accuracy & 96.5 & 89.7  \\ 
\hline
Joint goals & 85.5 & 80.9 \\ 
\hline
\end{tabular}
\caption{The test set accuracies \emph{overall} and for \emph{joint goals}  %and the three slots \emph{food, price range, area}
in the restaurant sub-domain.}
\label{tab:res2}
%\vspace*{-1em}o
\end{table}

The same model was trained on both the WOZ$2.0$ and the proposed MultiWOZ datasets, where the WOZ$2.0$ corpus consists of $1200$ single domain dialogues in the restaurant domain.
Although not directly comparable, Table \ref{tab:res2} shows that the performance of the model is consecutively poorer on the new dataset % for every slot 
compared to WOZ$2.0$. These results demonstrate how demanding is the new dataset as the conversations are richer and much longer. 
\begin{figure*}[t!]
\centering
 \includegraphics[width=.7\linewidth]{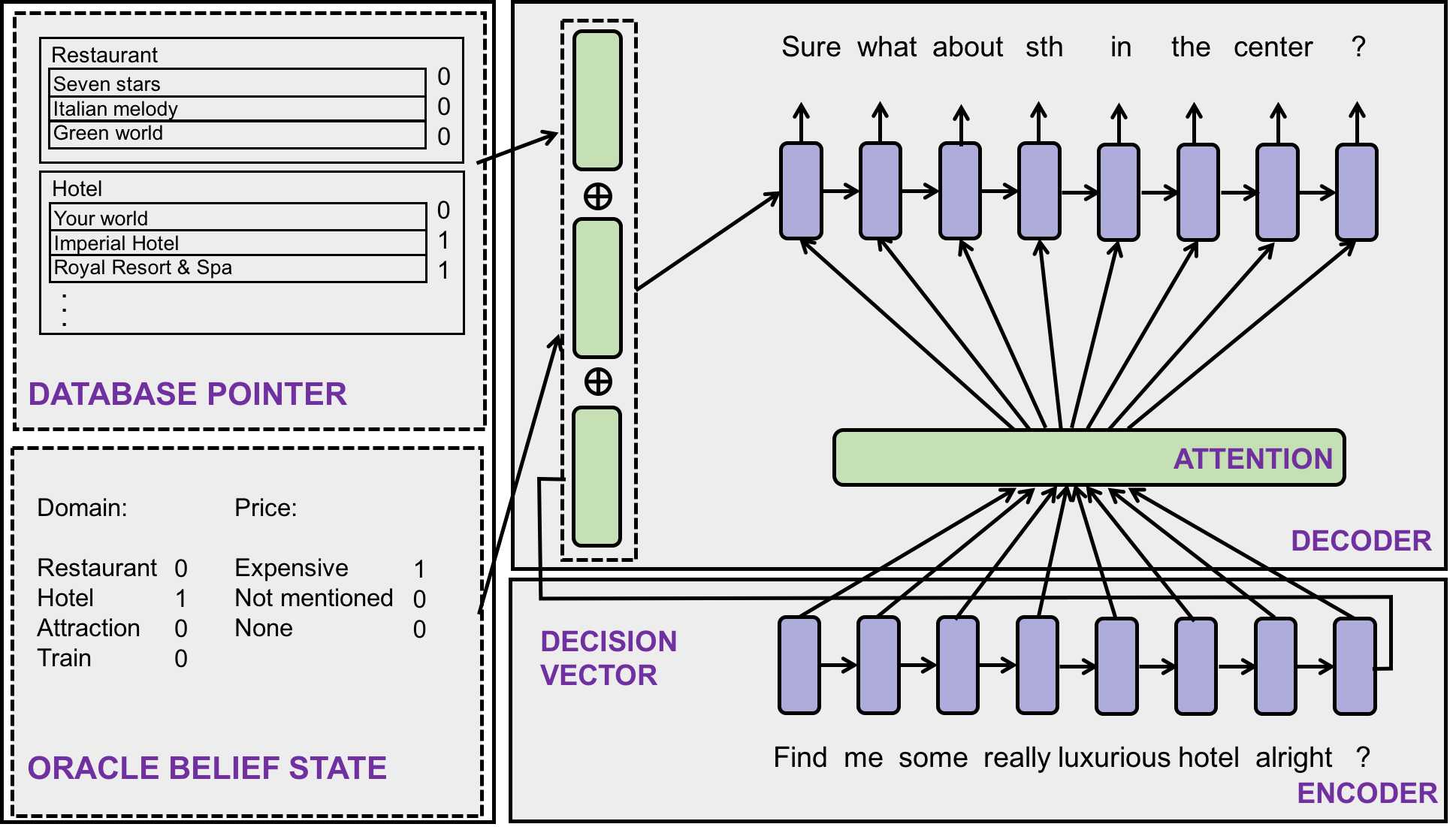}
 \caption{Architecture of the multi-domain response generator. The attention is conditioned on the oracle belief state and the database pointer.}
 \label{fig:n2n}
%\vspace*{-1em}
\end{figure*}

\subsection{Dialogue-Context-to-Text Generation}\label{sec:n2n}
After a robust dialogue state tracking module is built, the next challenge becomes the  dialogue management and response generation components.
These problems can either be addressed separately~\cite{POMDP-review}, or jointly in an end-to-end fashion~\cite{bordes2016learning,wen2016network,li2017end}.  In order to establish a clear benchmark where the performance of the composite of dialogue management and response generation is completely independent of the belief tracking, we experimented with a baseline neural response generation model with an \emph{oracle} belief-state obtained from the wizard annotations as discussed in Section \ref{sec:sysside}.\footnote{The model is publicly available at \url{https://github.com/budzianowski/multiwoz}} 

Following~\citet{wen2016network} which frames the dialogue as a context to response mapping problem, a sequence-to-sequence model~\citep{sutskever2014sequence} is augmented with a belief tracker and a discrete database accessing component as additional features to inform the word decisions in the decoder.
Note, in the original paper the belief tracker was pre-trained while in this work the annotations of the dialogue state are used as an oracle tracker. Figure \ref{fig:n2n} presents the architecture of the system \cite{budzianowski2018towards}.\\

{\noindent \bf Training and Evaluation} \hspace{1.5mm}
Since often times the evaluation of a dialogue system without a direct interaction with the real users can be misleading~\cite{liu2016not}, three different automatic metrics are included to ensure the result is better interpreted. Among them, the first two metrics relate to the dialogue task completion - whether the system has provided an appropriate entity (Inform rate) and then answered all the requested attributes (Success rate); while fluency is measured via BLEU score~\cite{papineni2002bleu}. The best models for both datasets were found through a grid search over a set of hyper-parameters such as the size of embeddings, learning rate and different recurrent architectures.

We trained the same neural architecture (taking into account different number of domains) on both MultiWOZ and Cam676 datasets. The best results on the Cam676 corpus were obtained with bidirectional GRU cell. In the case of MultiWOZ dataset, the LSTM cell serving as a decoder and an encoder achieved the highest score with the global type of attention \cite{bahdanau2014neural}. 
Table \ref{tab:e2e_eval} presents the results of a various of model architectures and shows several challenges. As expected, the model achieves almost perfect score on the Inform metric on the Cam676 dataset taking the advantage of an oracle belief state signal. However, even with the perfect dialogue state tracking of the user intent, the baseline models obtain almost $30\%$ lower score on the Inform metric on the new corpus. The addition of the attention improves the score on the Success metric on the new dataset by less than $1\%$. Nevertheless, as expected, the best model on MultiWOZ is still falling behind by a large margin in comparison to the results on the Cam676 corpus taking into account both Inform and Success metrics. As most of dialogues span over at least two domains, the model has to be much more effective in order to execute a successful dialogue. Moreover, the BLEU score on the MultiWOZ is lower than the one reported on the Cam676 dataset. This is mainly caused by the much more diverse linguistic expressions observed in the MultiWOZ dataset.

\begin{table*}[t!]
\begin{center}
\begin{tabular}{l|cc|cc}
  \multicolumn{3}{c|}{~~~~~~~~~~~~~~~~~~~~~~~~\textbf{Cam676}} & \multicolumn{2}{c}{\textbf{MultiWOZ}}  \\ & w/o attention & w/ attention & w/o attention & w/ attention \\\hline
Inform (\%)  & 99.17 &  99.58 & 71.29 & 71.33 \\
Success (\%) & 75.08 &73.75 & 60.29 & 60.96 \\
BLEU & 0.219 & 0.204 & 0.188 &  0.189 \\
\end{tabular}
\vspace*{-1em}
\end{center}
\caption{\label{tab:e2e_eval} Performance comparison of two different model architectures using a corpus-based evaluation.}
\vspace*{-1em}
\end{table*}

\subsection{Dialogue-Act-to-Text Generation}
Natural Language Generation from a structured meaning representation~\cite{oh2000stochastic,bohus2005sorry} has been a very popular research topic in the community, and the lack of data has been a long standing block for the field to adopt more machine learning methods.
Due to the additional annotation of the system acts, the MultiWOZ dataset serves as a new benchmark for studying natural language generation from a structured meaning representation.
In order to verify the difficulty of the collected dataset for the language generation task, we compare it to the SFX dataset~(see Table~\ref{fig:corpus_comparison}), which consists of around $5$k dialogue act and natural language sentence pairs.
We trained the same Semantically Conditioned Long Short-term Memory network (SC-LSTM) proposed by~\citet{wensclstm15} on both datasets and used the metrics as a proxy to estimate the difficulty of the two corpora. 
To make a fair comparison, we constrained our dataset to only the restaurant sub-domain which contains around $25$k dialogue turns.
To give more statistics about the two datasets: the SFX corpus has $9$ different act types with $12$ slots  comparing to $12$ acts and $14$ slots in our corpus.
The best model for both datasets was found through a grid search over a set of hyper-parameters such as the size of embeddings, learning rate, and number of LSTM layers.\footnote{The model is publicly available at \url{https://github.com/andy194673/nlg-sclstm-multiwoz}}

Table \ref{tab:acts} presents the results on two metrics: BLEU score~\cite{papineni2002bleu} and slot error rate (SER)~\cite{wensclstm15}.
The significantly lower metrics on the MultiWOZ corpus showed that it is much more challenging than the SFX restaurant dataset.
This is probably due to the fact that more than $60\%$ of the dialogue turns are composed of at least two system acts, which greatly harms the performance of the existing model.

\begin{table}[h!]
\centering
\begin{tabular}{ |c|c|c|}
\hline
\multirow{ 2}{*}{\textbf{Metric}} & \multirow{ 2}{*}{\textbf{SFX}} & \textbf{MultiWOZ}  \\
& & \textbf{(restaurant)}\\
 \hline
 SER (\%) & 0.46 &  4.378\\ 
\hline
BLEU & 0.731 & 0.616 \\ 
\hline
\end{tabular}
\caption{The test set slot error rate (SER) and BLEU on the SFX dataset and the MultiWOZ restaurant subset. }\label{tab:acts}
\vspace*{-1em}
\end{table}

\begin{table}[h!]
\centering
\begin{tabular}{ |c|c|c|}
\hline
\multirow{ 2}{*}{} & \textbf{Single} & \textbf{Multi}  \\
 \hline
 \# of dialogues & 3,406 &  7,032\\ 
\hline
\# of domains & 1-2 & 2-6\\ 
\hline
\end{tabular}
\caption{The test set slot error rate (SER) and BLEU on the SFX dataset and the MultiWOZ restaurant subset. }\label{tab:acts}
\vspace*{-1em}
\end{table}

\section{Conclusions}
As more and more speech oriented applications are commercially deployed, the necessity of building an entirely data-driven conversational agent becomes more apparent. Various corpora were gathered to enable data-driven approaches to dialogue modelling. 
To date, however, the available datasets were usually constrained in linguistic variability or lacking multi-domain use cases. In this paper, we established a data-collection pipeline entirely based on crowd-sourcing enabling to gather a large scale, linguistically rich corpus of human-human conversations.
We hope that MultiWOZ offers valuable training data and a new challenging testbed for existing modular-based approaches ranging from belief tracking to dialogue acts generation.
Moreover, the scale of the data should help push forward research in the end-to-end dialogue modelling.

\section*{Acknowledgments}
This work was funded by a Google Faculty Research Award (RG91111), an EPSRC studentship (RG80792), an EPSRC grant (EP/M018946/1) and by Toshiba Research Europe Ltd, Cambridge  Research  Laboratory (RG85875). The authors thank many excellent Mechanical Turk contributors for building this dataset. The authors would also like to thank Thang Minh Luong for his support for this project and Nikola Mrk\v{s}i\'{c} and anonymous reviewers for their constructive feedback. The data is available at \url{https://github.com/budzianowski/multiwoz}.
\newpage
\bibliography{emnlp2018}

\begin{thebibliography}{52}
\expandafter\ifx\csname natexlab\endcsname\relax\def\natexlab#1{#1}\fi

\bibitem[{Asri et~al.(2017)Asri, Schulz, Sharma, Zumer, Harris, Fine, Mehrotra,
  and Suleman}]{asri2017frames}
Layla~El Asri, Hannes Schulz, Shikhar Sharma, Jeremie Zumer, Justin Harris,
  Emery Fine, Rahul Mehrotra, and Kaheer Suleman. 2017.
\newblock Frames: A corpus for adding memory to goal-oriented dialogue systems.
\newblock \emph{Proceedings of SigDial}.

\bibitem[{Bahdanau et~al.(2014)Bahdanau, Cho, and Bengio}]{bahdanau2014neural}
Dzmitry Bahdanau, Kyunghyun Cho, and Yoshua Bengio. 2014.
\newblock Neural machine translation by jointly learning to align and
  translate.
\newblock \emph{ICLR}.

\bibitem[{Black et~al.(2011)Black, Burger, Conkie, Hastie, Keizer, Lemon,
  Merigaud, Parent, Schubiner, Thomson et~al.}]{black2011spoken}
Alan~W Black, Susanne Burger, Alistair Conkie, Helen Hastie, Simon Keizer,
  Oliver Lemon, Nicolas Merigaud, Gabriel Parent, Gabriel Schubiner, Blaise
  Thomson, et~al. 2011.
\newblock Spoken dialog challenge 2010: Comparison of live and control test
  results.
\newblock In \emph{Proceedings of the SIGDIAL 2011 Conference}, pages 2--7.
  Association for Computational Linguistics.

\bibitem[{Bohus and Rudnicky(2005)}]{bohus2005sorry}
Dan Bohus and Alexander~I Rudnicky. 2005.
\newblock Sorry, i didn't catch that! - an investigation of non-understanding
  errors and recovery strategies.
\newblock In \emph{6th SIGdial workshop on discourse and dialogue}.

\bibitem[{Bordes et~al.(2017)Bordes, Boureau, and Weston}]{bordes2016learning}
Antoine Bordes, Y-Lan Boureau, and Jason Weston. 2017.
\newblock Learning end-to-end goal-oriented dialog.
\newblock \emph{Proceedings of ICLR}.

\bibitem[{Budzianowski et~al.(2018)Budzianowski, Casanueva, Tseng, and Ga{\v
  s}i{\'c}}]{budzianowski2018towards}
Pawe{\l} Budzianowski, I{\~n}igo Casanueva, Bo-Hsiang Tseng, and Milica Ga{\v
  s}i{\'c}. 2018.
\newblock Towards end-to-end multi-domain dialogue modelling.
\newblock \emph{Tech. Rep. CUED/F-INFENG/TR.706, University of Cambridge,
  Engineering Department}.

\bibitem[{Bunt(2006)}]{bunt2006dimensions}
Harry Bunt. 2006.
\newblock Dimensions in dialogue act annotation.
\newblock In \emph{Proc. of LREC}, volume~6, pages 919--924.

\bibitem[{Eric et~al.(2017)Eric, Krishnan, Charette, and Manning}]{eric2017key}
Mihail Eric, Lakshmi Krishnan, Francois Charette, and Christopher~D Manning.
  2017.
\newblock Key-value retrieval networks for task-oriented dialogue.
\newblock In \emph{Proceedings of the 18th Annual SIGdial Meeting on Discourse
  and Dialogue}, pages 37--49.

\bibitem[{Fleiss(1971)}]{fleiss1971measuring}
Joseph~L Fleiss. 1971.
\newblock Measuring nominal scale agreement among many raters.
\newblock \emph{Psychological bulletin}, 76(5):378.

\bibitem[{Ga{\v{s}}i{\'c} et~al.(2014)Ga{\v{s}}i{\'c}, Kim, Tsiakoulis,
  Breslin, Henderson, Szummer, Thomson, and Young}]{gavsic2014incremental}
Milica Ga{\v{s}}i{\'c}, Dongho Kim, Pirros Tsiakoulis, Catherine Breslin,
  Matthew Henderson, Martin Szummer, Blaise Thomson, and Steve Young. 2014.
\newblock Incremental on-line adaptation of pomdp-based dialogue managers to
  extended domains.
\newblock In \emph{Interspeech}.

\bibitem[{Ga{\v s}i{\'c} and Young(2014)}]{GPRL}
Milica Ga{\v s}i{\'c} and Steve Young. 2014.
\newblock Gaussian processes for pomdp-based dialogue manager optimization.
\newblock \emph{TASLP}, 22(1):28--40.

\bibitem[{Hemphill et~al.(1990)Hemphill, Godfrey, and
  Doddington}]{hemphill1990atis}
Charles~T Hemphill, John~J Godfrey, and George~R Doddington. 1990.
\newblock The atis spoken language systems pilot corpus.
\newblock In \emph{Speech and Natural Language: Proceedings of a Workshop Held
  at Hidden Valley, Pennsylvania}.

\bibitem[{Henderson et~al.(2014{\natexlab{a}})Henderson, Thomson, and
  Williams}]{Henderson2014a}
M.~Henderson, B.~Thomson, and J.~Williams. 2014{\natexlab{a}}.
\newblock The second dialog state tracking challenge.
\newblock In \emph{Proceedings of SIGdial}.

\bibitem[{Henderson et~al.(2014{\natexlab{b}})Henderson, Thomson, and
  Young}]{Henderson2014b}
M.~Henderson, B.~Thomson, and S.~J. Young. 2014{\natexlab{b}}.
\newblock {Word-based Dialog State Tracking with Recurrent Neural Networks}.
\newblock In \emph{Proceedings of SIGdial}.

\bibitem[{Henderson et~al.(2014{\natexlab{c}})Henderson, Thomson, and
  Williams}]{henderson2014third}
Matthew Henderson, Blaise Thomson, and Jason~D Williams. 2014{\natexlab{c}}.
\newblock The third dialog state tracking challenge.
\newblock In \emph{Spoken Language Technology Workshop (SLT), 2014 IEEE}, pages
  324--329. IEEE.

\bibitem[{Henderson et~al.(2013)Henderson, Thomson, and
  Young}]{henderson2013deep}
Matthew Henderson, Blaise Thomson, and Steve Young. 2013.
\newblock Deep neural network approach for the dialog state tracking challenge.
\newblock In \emph{Proceedings of the SIGDIAL 2013 Conference}, pages 467--471.

\bibitem[{Kelley(1984)}]{kelley1984iterative}
John~F Kelley. 1984.
\newblock An iterative design methodology for user-friendly natural language
  office information applications.
\newblock \emph{ACM Transactions on Information Systems (TOIS)}, 2(1):26--41.

\bibitem[{Kiddon et~al.(2016)Kiddon, Zettlemoyer, and
  Choi}]{kiddon2016globally}
Chlo{\'e} Kiddon, Luke Zettlemoyer, and Yejin Choi. 2016.
\newblock Globally coherent text generation with neural checklist models.
\newblock In \emph{Proceedings of the 2016 Conference on Empirical Methods in
  Natural Language Processing}, pages 329--339.

\bibitem[{Kim et~al.(2016)Kim, D'Haro, Banchs, Williams, Henderson, and
  Yoshino}]{kim2016fifth}
Seokhwan Kim, Luis~Fernando D'Haro, Rafael~E Banchs, Jason~D Williams, Matthew
  Henderson, and Koichiro Yoshino. 2016.
\newblock The fifth dialog state tracking challenge.
\newblock In \emph{Spoken Language Technology Workshop (SLT), 2016 IEEE}, pages
  511--517. IEEE.

\bibitem[{Kim et~al.(2017)Kim, D’Haro, Banchs, Williams, and
  Henderson}]{kim2017fourth}
Seokhwan Kim, Luis~Fernando D’Haro, Rafael~E Banchs, Jason~D Williams, and
  Matthew Henderson. 2017.
\newblock The fourth dialog state tracking challenge.
\newblock In \emph{Dialogues with Social Robots}, pages 435--449. Springer.

\bibitem[{Li et~al.(2016)Li, Galley, Brockett, Gao, and Dolan}]{LiGBGD15}
Jiwei Li, Michel Galley, Chris Brockett, Jianfeng Gao, and Bill Dolan. 2016.
\newblock A diversity-promoting objective function for neural conversation
  models.
\newblock In \emph{NAACL-HLT}, pages 110--119, San Diego, California.
  Association for Computational Linguistics.

\bibitem[{Li et~al.(2017)Li, Chen, Li, Gao, and Celikyilmaz}]{li2017end}
Xiujun Li, Yun-Nung Chen, Lihong Li, Jianfeng Gao, and Asli Celikyilmaz. 2017.
\newblock End-to-end task-completion neural dialogue systems.
\newblock In \emph{Proceedings of the Eighth International Joint Conference on
  Natural Language Processing (Volume 1: Long Papers)}, volume~1, pages
  733--743.

\bibitem[{Liu et~al.(2016)Liu, Lowe, Serban, Noseworthy, Charlin, and
  Pineau}]{liu2016not}
Chia-Wei Liu, Ryan Lowe, Iulian Serban, Mike Noseworthy, Laurent Charlin, and
  Joelle Pineau. 2016.
\newblock How not to evaluate your dialogue system: An empirical study of
  unsupervised evaluation metrics for dialogue response generation.
\newblock In \emph{Proceedings of the 2016 Conference on Empirical Methods in
  Natural Language Processing}, pages 2122--2132.

\bibitem[{Lowe et~al.(2015)Lowe, Pow, Serban, and Pineau}]{lowe2015ubuntu}
Ryan Lowe, Nissan Pow, Iulian~V Serban, and Joelle Pineau. 2015.
\newblock The ubuntu dialogue corpus: A large dataset for research in
  unstructured multi-turn dialogue systems.
\newblock In \emph{16th Annual Meeting of the Special Interest Group on
  Discourse and Dialogue}, page 285.

\bibitem[{McCarthy et~al.(1955)McCarthy, Minsky, Rochester, and
  Shannon}]{dartmouth1955}
J.~McCarthy, M.~L. Minsky, N.~Rochester, and C.~E. Shannon. 1955.
\newblock A proposal for the dartmouth summer research project on artificial
  intelligence.

\bibitem[{Mesnil et~al.(2015)Mesnil, Dauphin, Yao, Bengio, Deng, Hakkani-Tur,
  He, Heck, Tur, Yu et~al.}]{mesnil2015using}
Gr{\'e}goire Mesnil, Yann Dauphin, Kaisheng Yao, Yoshua Bengio, Li~Deng, Dilek
  Hakkani-Tur, Xiaodong He, Larry Heck, Gokhan Tur, Dong Yu, et~al. 2015.
\newblock Using recurrent neural networks for slot filling in spoken language
  understanding.
\newblock \emph{IEEE/ACM Transactions on Audio, Speech, and Language
  Processing}, 23(3):530--539.

\bibitem[{Mrk{\v{s}}i{\'c} et~al.(2017{\natexlab{a}})Mrk{\v{s}}i{\'c},
  S{\'e}aghdha, Wen, Thomson, and Young}]{mrkvsic2017neural}
Nikola Mrk{\v{s}}i{\'c}, Diarmuid~{\'O} S{\'e}aghdha, Tsung-Hsien Wen, Blaise
  Thomson, and Steve Young. 2017{\natexlab{a}}.
\newblock Neural belief tracker: Data-driven dialogue state tracking.
\newblock In \emph{Proceedings of the 55th Annual Meeting of the Association
  for Computational Linguistics (Volume 1: Long Papers)}, volume~1, pages
  1777--1788.

\bibitem[{Mrk{\v{s}}i{\'c} et~al.(2017{\natexlab{b}})Mrk{\v{s}}i{\'c},
  Vuli{\'c}, S{\'e}aghdha, Leviant, Reichart, Ga{\v{s}}i{\'c}, Korhonen, and
  Young}]{mrkvsic2017semantic}
Nikola Mrk{\v{s}}i{\'c}, Ivan Vuli{\'c}, Diarmuid~{\'O} S{\'e}aghdha, Ira
  Leviant, Roi Reichart, Milica Ga{\v{s}}i{\'c}, Anna Korhonen, and Steve
  Young. 2017{\natexlab{b}}.
\newblock Semantic specialization of distributional word vector spaces using
  monolingual and cross-lingual constraints.
\newblock \emph{Transactions of the Association of Computational Linguistics},
  5(1):309--324.

\bibitem[{Oh and Rudnicky(2000)}]{oh2000stochastic}
Alice~H Oh and Alexander~I Rudnicky. 2000.
\newblock Stochastic language generation for spoken dialogue systems.
\newblock In \emph{Proceedings of the 2000 ANLP/NAACL Workshop on
  Conversational systems-Volume 3}, pages 27--32. Association for Computational
  Linguistics.

\bibitem[{Paek and Pieraccini(2008)}]{paek2008automating}
Tim Paek and Roberto Pieraccini. 2008.
\newblock Automating spoken dialogue management design using machine learning:
  An industry perspective.
\newblock \emph{Speech communication}, 50(8-9):716--729.

\bibitem[{Papineni et~al.(2002)Papineni, Roukos, Ward, and
  Zhu}]{papineni2002bleu}
Kishore Papineni, Salim Roukos, Todd Ward, and Wei-Jing Zhu. 2002.
\newblock Bleu: a method for automatic evaluation of machine translation.
\newblock In \emph{Proceedings of the 40th annual meeting on association for
  computational linguistics}, pages 311--318. Association for Computational
  Linguistics.

\bibitem[{Ram et~al.(2018)Ram, Prasad, Khatri, Venkatesh, Gabriel, Liu, Nunn,
  Hedayatnia, Cheng, Nagar et~al.}]{ram2018conversational}
Ashwin Ram, Rohit Prasad, Chandra Khatri, Anu Venkatesh, Raefer Gabriel, Qing
  Liu, Jeff Nunn, Behnam Hedayatnia, Ming Cheng, Ashish Nagar, et~al. 2018.
\newblock Conversational ai: The science behind the alexa prize.
\newblock \emph{arXiv preprint arXiv:1801.03604}.

\bibitem[{Ramadan et~al.(2018)Ramadan, Budzianowski, and
  Ga\v{s}i\'{c}}]{ramadan2018large}
Osman Ramadan, Pawe{\l} Budzianowski, and Milica Ga\v{s}i\'{c}. 2018.
\newblock Large-scale multi-domain belief tracking with knowledge sharing.
\newblock In \emph{Proceedings of the 56th Annual Meeting of the Association
  for Computational Linguistics}, volume~2, pages 432--437.

\bibitem[{Rastogi et~al.(2017)Rastogi, Hakkani-Tur, and Heck}]{rast:17}
Abhinav Rastogi, Dilek Hakkani-Tur, and Larry Heck. 2017.
\newblock Scalable multi-domain dialogue state tracking.
\newblock \emph{arXiv preprint arXiv:1712.10224}.

\bibitem[{Raux et~al.(2005)Raux, Langner, Bohus, Black, and
  Eskenazi}]{raux2005let}
Antoine Raux, Brian Langner, Dan Bohus, Alan~W Black, and Maxine Eskenazi.
  2005.
\newblock Let's go public! taking a spoken dialog system to the real world.
\newblock In \emph{Ninth European Conference on Speech Communication and
  Technology}.

\bibitem[{Ritter et~al.(2010)Ritter, Cherry, and
  Dolan}]{ritter2010unsupervised}
Alan Ritter, Colin Cherry, and Bill Dolan. 2010.
\newblock Unsupervised modeling of twitter conversations.
\newblock In \emph{Human Language Technologies: The 2010 Annual Conference of
  the North American Chapter of the Association for Computational Linguistics},
  pages 172--180.

\bibitem[{Schrading et~al.(2015)Schrading, Alm, Ptucha, and
  Homan}]{schrading2015analysis}
Nicolas Schrading, Cecilia~Ovesdotter Alm, Ray Ptucha, and Christopher Homan.
  2015.
\newblock An analysis of domestic abuse discourse on reddit.
\newblock In \emph{Proceedings of the 2015 Conference on Empirical Methods in
  Natural Language Processing}, pages 2577--2583.

\bibitem[{Seneff and Polifroni(2000)}]{Seneff2000}
Stephanie Seneff and Joseph Polifroni. 2000.
\newblock Dialogue management in the mercury flight reservation system.
\newblock In \emph{Proceedings of the 2000 ANLP/NAACL Workshop on
  Conversational Systems - Volume 3}, ANLP/NAACL-ConvSyst '00, pages 11--16,
  Stroudsburg, PA, USA. Association for Computational Linguistics.

\bibitem[{Shah et~al.(2018)Shah, Hakkani-Tur, Tur, Rastogi, Bapna, Nayak, and
  Heck}]{shah2018building}
P~Shah, D~Hakkani-Tur, G~Tur, A~Rastogi, A~Bapna, N~Nayak, and L~Heck. 2018.
\newblock Building a conversational agent overnight with dialogue self-play.
\newblock \emph{arXiv preprint arXiv:1801.04871}.

\bibitem[{Stent et~al.(2005)Stent, Marge, and Singhai}]{stent2005evaluating}
Amanda Stent, Matthew Marge, and Mohit Singhai. 2005.
\newblock Evaluating evaluation methods for generation in the presence of
  variation.
\newblock In \emph{International Conference on Intelligent Text Processing and
  Computational Linguistics}, pages 341--351. Springer.

\bibitem[{Sutskever et~al.(2014)Sutskever, Vinyals, and
  Le}]{sutskever2014sequence}
Ilya Sutskever, Oriol Vinyals, and Quoc~V Le. 2014.
\newblock Sequence to sequence learning with neural networks.
\newblock In \emph{Advances in neural information processing systems}, pages
  3104--3112.

\bibitem[{Tegho et~al.(2018)Tegho, Budzianowski, and Ga{\v
  s}i{\'c}}]{tegho2018benchmarking}
Christopher Tegho, Pawe{\l} Budzianowski, and Milica Ga{\v s}i{\'c}. 2018.
\newblock Benchmarking uncertainty estimates with deep reinforcement learning
  for dialogue policy optimisation.
\newblock In \emph{IEEE ICASSP 2018}.

\bibitem[{Traum(1999)}]{Traum1999}
David~R. Traum. 1999.
\newblock \emph{Foundations of Rational Agency}, chapter Speech Acts for
  Dialogue Agents. Springer.

\bibitem[{Traum and Hinkelman(1992)}]{traum1992conversation}
David~R Traum and Elizabeth~A Hinkelman. 1992.
\newblock Conversation acts in task-oriented spoken dialogue.
\newblock \emph{Computational intelligence}, 8(3):575--599.

\bibitem[{Traum and Larsson(2003)}]{traum2003information}
David~R Traum and Staffan Larsson. 2003.
\newblock The information state approach to dialogue management.
\newblock In \emph{Current and new directions in discourse and dialogue}, pages
  325--353. Springer.

\bibitem[{Vinyals and Le(2015)}]{vinyals2015neural}
Oriol Vinyals and Quoc Le. 2015.
\newblock A neural conversational model.
\newblock \emph{arXiv preprint arXiv:1506.05869}.

\bibitem[{Wen et~al.(2016)Wen, Ga{\v s}i{\'c}, Mrksic, Rojas-Barahona, Su,
  Vandyke, and Young}]{wen2016multi}
Tsung-Hsien Wen, Milica Ga{\v s}i{\'c}, Nikola Mrksic, Lina~M Rojas-Barahona,
  Pei-Hao Su, David Vandyke, and Steve Young. 2016.
\newblock Multi-domain neural network language generation for spoken dialogue
  systems.
\newblock \emph{ACL}.

\bibitem[{Wen et~al.(2015)Wen, Ga{\v{s}}i\'c, Mrk{\v{s}}i\'c, Su, Vandyke, and
  Young}]{wensclstm15}
Tsung-Hsien Wen, Milica Ga{\v{s}}i\'c, Nikola Mrk{\v{s}}i\'c, Pei-Hao Su, David
  Vandyke, and Steve Young. 2015.
\newblock Semantically conditioned lstm-based natural language generation for
  spoken dialogue systems.
\newblock In \emph{Proceedings of the 2015 Conference on Empirical Methods in
  Natural Language Processing (EMNLP)}.

\bibitem[{Wen et~al.(2017)Wen, Vandyke, Mrksic, Ga{\v s}i{\'c}, Rojas-Barahona,
  Su, Ultes, and Young}]{wen2016network}
Tsung-Hsien Wen, David Vandyke, Nikola Mrksic, Milica Ga{\v s}i{\'c}, Lina~M
  Rojas-Barahona, Pei-Hao Su, Stefan Ultes, and Steve Young. 2017.
\newblock A network-based end-to-end trainable task-oriented dialogue system.
\newblock \emph{EACL}.

\bibitem[{Williams et~al.(2013)Williams, Raux, Ramachandran, and
  Black}]{williams2013dialog}
Jason Williams, Antoine Raux, Deepak Ramachandran, and Alan Black. 2013.
\newblock The dialog state tracking challenge.
\newblock In \emph{Proceedings of the SIGDIAL 2013 Conference}, pages 404--413.

\bibitem[{Young et~al.(2013)Young, Ga{\v s}i\'{c}, Thomson, and
  Williams}]{POMDP-review}
Steve Young, Milica Ga{\v s}i\'{c}, Blaise Thomson, and Jason Williams. 2013.
\newblock {POMDP-based Statistical Spoken Dialogue Systems: a Review}.
\newblock In \emph{Proc of IEEE}, volume~99, pages 1--20.

\bibitem[{Zhao and Eskenazi(2016)}]{zhao2016towards}
Tiancheng Zhao and Maxine Eskenazi. 2016.
\newblock Towards end-to-end learning for dialog state tracking and management
  using deep reinforcement learning.
\newblock In \emph{17th Annual Meeting of the Special Interest Group on
  Discourse and Dialogue}, page~1.

\end{thebibliography}
\bibliographystyle{acl_natbib_nourl}
\clearpage

\newpage
\appendix
\renewcommand{\thefigure}{A\arabic{figure}}

\setcounter{figure}{0}
\section{MTurk Website Set-up}
\label{sec:website}
\begin{figure*}
\centering
\includegraphics[width=.7\linewidth]{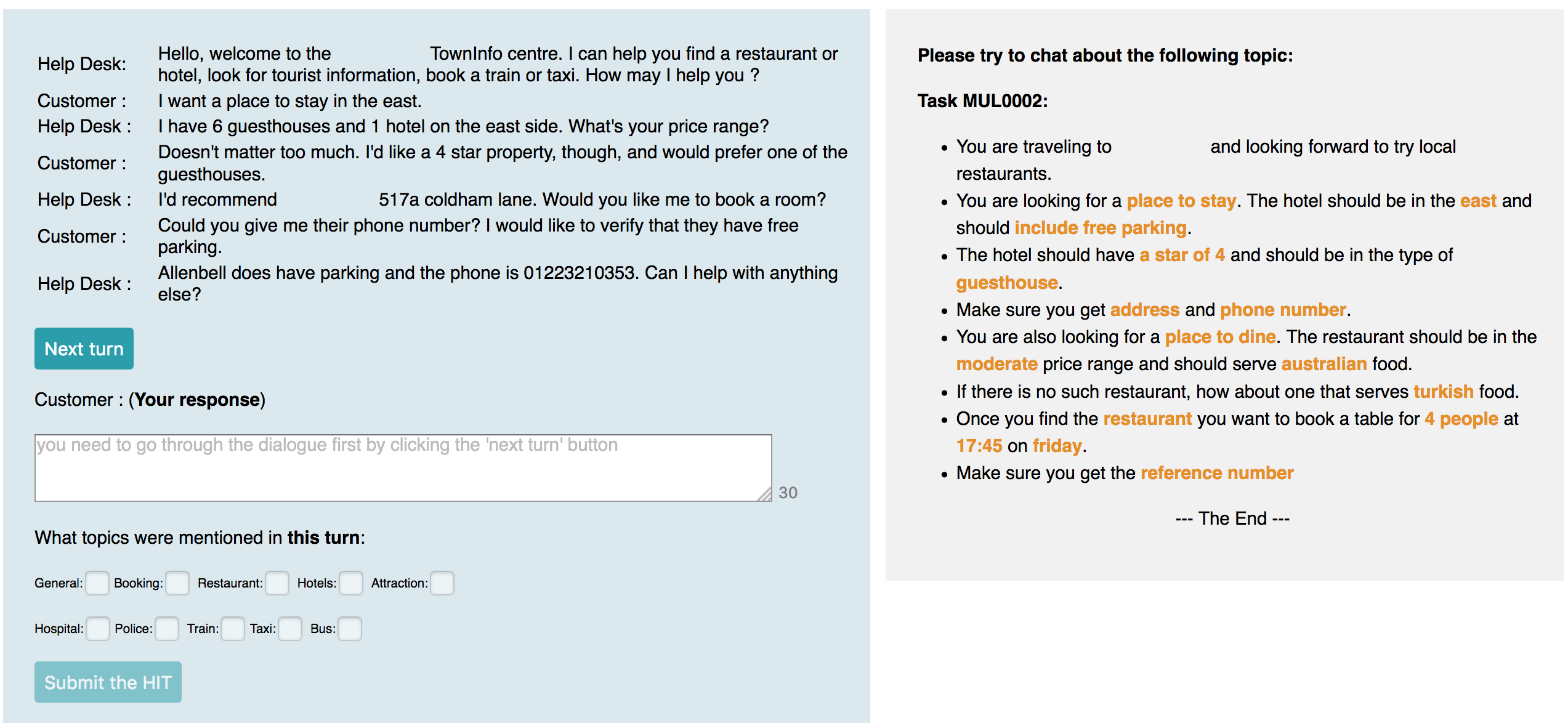}
  \caption{Interface from the User side }
  \label{fig:user_side}
\end{figure*}

\begin{figure*}
\centering
 \includegraphics[width=.7\linewidth]{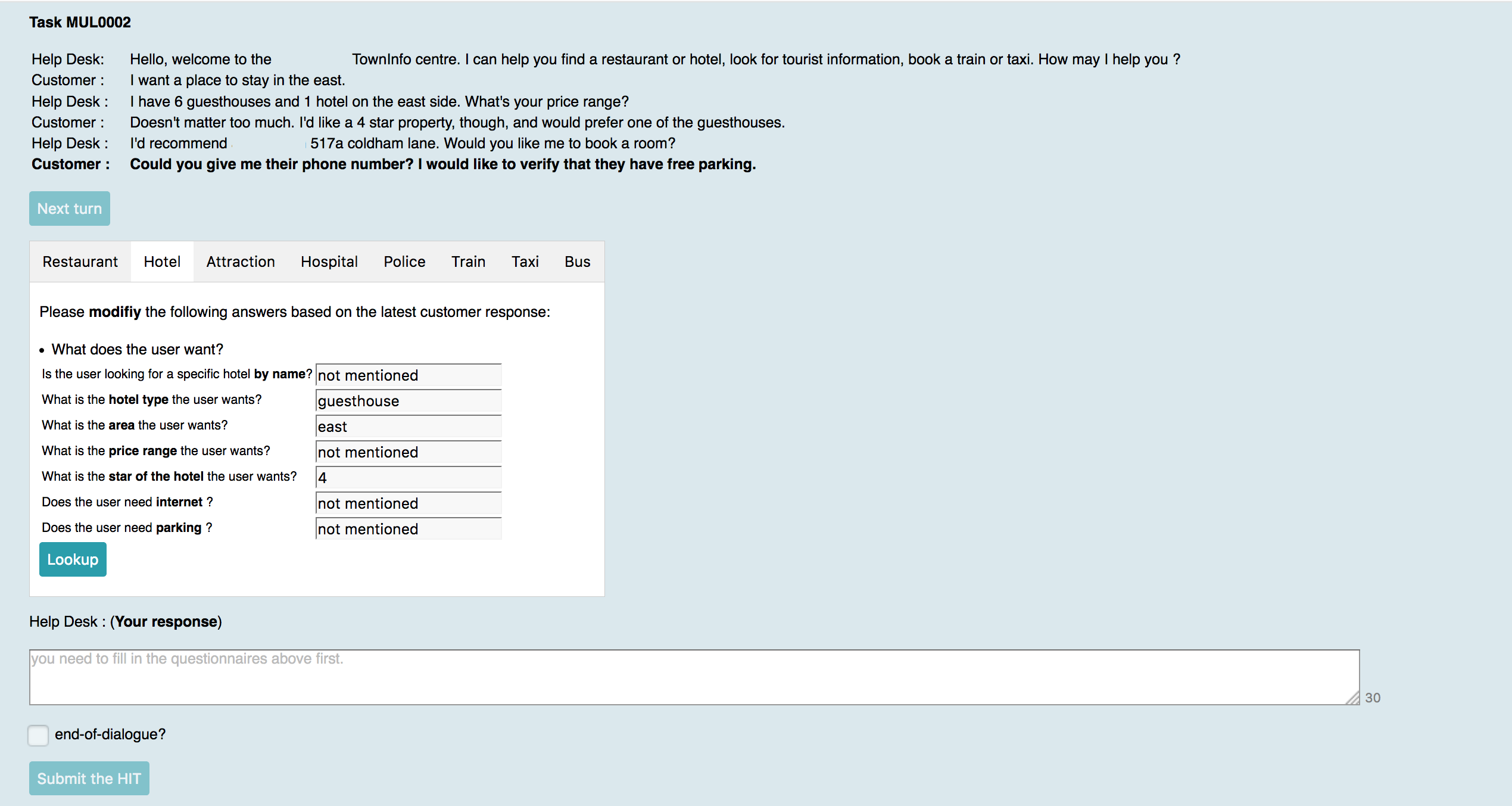}
  \caption{Interface from the Wizard side }
  \label{fig:system_side}
\end{figure*}

\begin{figure*}
\centering
 \includegraphics[width=.6\linewidth]{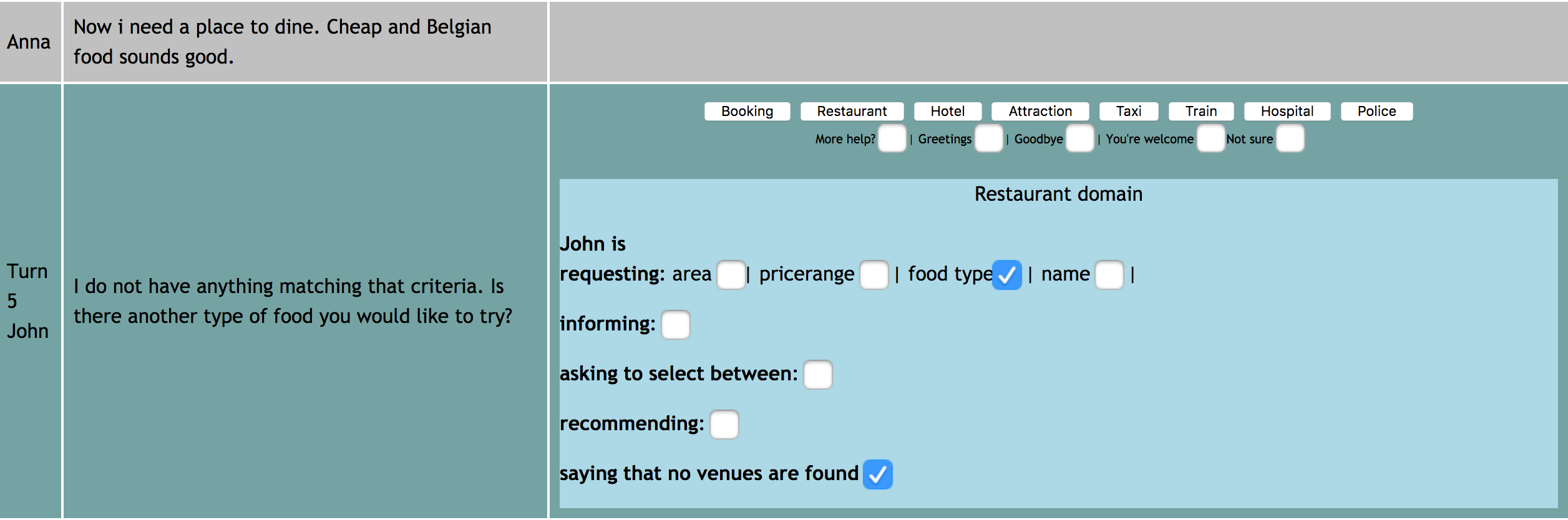}
  \caption{Interface for the annotation., }
  \label{fig:acts_side}
\end{figure*}

Figure \ref{fig:user_side} presents the user side interface where the worker needs to properly respond given the task description and the dialogue history. Figure \ref{fig:system_side} shows the wizard page with the GUI over all domains. Finally, Figure \ref{fig:acts_side} shows the set-up for annotation of the system acts with Restaurant domain being turned on.

\newpage

\end{document}